\pdfoutput=1

\documentclass[11pt]{article}

\usepackage[final]{acl}

\usepackage{times}
\usepackage{latexsym}
\usepackage{enumitem}
\usepackage{longtable}
\usepackage{supertabular}
\usepackage[T1]{fontenc}

\usepackage[utf8]{inputenc}

\usepackage{microtype}

\usepackage{inconsolata}

\usepackage{url}
\usepackage{graphicx} 
\usepackage{subfigure}
\usepackage{algpseudocode}
\usepackage{algorithm}
\usepackage{amsmath}

\usepackage{pdfpages}  

\graphicspath{{./images/}}

\title{OmAgent: A Multi-modal Agent Framework for Complex Video Understanding with Task Divide-and-Conquer}

\author{
 \textbf{Lu Zhang\textsuperscript{1}},
 \textbf{Tiancheng Zhao\textsuperscript{1,2}},
 \textbf{Heting Ying\textsuperscript{1}},
 \textbf{Yibo Ma\textsuperscript{1}},
 \textbf{Kyusong Lee\textsuperscript{1,2}}
\\
\\
 \textsuperscript{1}Om AI Research,
 \textsuperscript{2}Binjiang Institute of Zhejiang University
\\
 \small{
   \texttt{\{zhang\_lu, ying\_heting, ma\_yibo\}@hzlh.com}
 }
\
 \small{
   \texttt{\{tianchez, kyusongl\}@zju-bj.com}
 }
}

\begin{document}
\maketitle
\begin{abstract}
Recent advancements in Large Language Models (LLMs) have expanded their capabilities to multimodal contexts, including comprehensive video understanding. However, processing extensive videos such as 24-hour CCTV footage or full-length films presents significant challenges due to the vast data and processing demands. Traditional methods, like extracting key frames or converting frames to text, often result in substantial information loss. To address these shortcomings, we develop OmAgent, efficiently stores and retrieves relevant video frames for specific queries, preserving the detailed content of videos. Additionally, it features an Divide-and-Conquer Loop capable of autonomous reasoning, dynamically invoking APIs and tools to enhance query processing and accuracy. This approach ensures robust video understanding, significantly reducing information loss. Experimental results affirm OmAgent's efficacy in handling various types of videos and complex tasks. Moreover, we have endowed it with greater autonomy and a robust tool-calling system, enabling it to accomplish even more intricate tasks. Code: \url{https://github.com/om-ai-lab/OmAgent}
\end{abstract}

\section{Introduction}
\label{sec:intro}

Large Language Models (LLMs) have advanced remarkably in recent years, greatly expanding their capabilities across various applications~\cite{touvron2023llama,touvron2023llama2,qwen,GPT-4}. As these models have evolved, they have been increasingly applied to multimodal contexts, allowing them to process and interpret not just text but also images and other media types. Initially, the focus was on single images, utilizing the models' ability to generate and understand detailed descriptions or responses based on static visuals~\cite{awadalla2023openflamingo,liu2023llava,liu2024llavanext,GPT-4(vision)}

However, as LLMs have become more complex and powerful, there has been growing interest in applying them to more dynamic media, such as video content~\cite{lin2023video, zhu2023languagebind,zhang2023video,zhao2023learning,tang2023video}. This interest arises from the potential to provide deeper and more nuanced interpretations of video data, similar to how humans understand and interact with moving images and sound. Currently, most video understanding models are limited to processing short videos, typically only a few minutes or even several seconds long. Despite these advancements, significant challenges remain, especially in handling long video inputs like 24-hour CCTV footage or full-length movies, which involve massive amounts of data and require substantial processing power.

Traditionally, one solution has been to extract key frames from these long videos or convert all frames into textual descriptions before processing~\cite{lin2023mmvid,yang2023mmreact}. While this approach makes the task more manageable for LLMs, it often results in information loss. Key frame extraction might miss subtle but important details in the omitted frames, and converting visual data to text can oversimplify or misrepresent visual nuances, leading to a less accurate understanding of the content.

To address these limitations, the application of Retrieval-Augmented Generation (RAG) technology in video understanding has emerged as a promising solution~\cite{chen2017reading, Liu_LlamaIndex_2022,Chase_LangChain_2022,arefeen2024irag}. RAG enables the storage and efficient retrieval of video frames based on their relevance to a query. This method allows for more precise and contextual responses by referencing specific content directly from the video, rather than relying on potentially incomplete or inaccurate textual summaries. However, given that video is a continuous stream of information, when it is stored, this continuous flow is segmented into discrete blocks of data. The information lost during this segmentation is irretrievable.

To address the aforementioned issues, we attempt to analyze the human approach to handling complex, long video question-answering tasks, seeking breakthroughs from this perspective. When a person watches a content-rich, lengthy video, such as a movie, they retain a general impression of the movie in their mind. This impression includes a rough outline of the video's content at various time points. When asked about specific details, the person may not recall the details immediately but can quickly locate the relevant time point in the video and rewatch the segment to retrieve the missing information. The \textbf{key insight} of OmAgent is to replicate this process by integrating multimodal RAG and generalist AI agent. OmAgent consists of two main components: \textbf{(1)} A video2RAG video preprocessor to extract and store the generalized information from the video, akin to the foundational impression a video imprints upon the viewer's memory. \textbf{(2)} A Divide-and-Conquer Loop (DnC Loop) for task planning and execution which equipped with tool invocation capabilities. 

We abstract the human ability to reposition and review video details as a tool named "rewinder," which can be autonomously selected and utilized by an AI agent, similar to how a person might use a video player's progress bar to navigate to points of interest. OmAgent not only can retrieve detailed information from videos but also can actively seek external information, enabling more advanced video understanding and question-answering. Existing benchmarks are insufficient to accurately quantify these capabilities, so we propose a new complex video understanding benchmark to fulfill this task. The contributions are:
    \textbf{(1)} OmAgent, the first complex video understanding framework integrating multimodal RAG and generalist AI agent.
    \textbf{(2)} A benchmark dataset that contains 2000+ Q\&A pairs for evaluating video understanding systems.
    \textbf{(3)} Experiments that shows the proposed agentic method is able to outperform strong baselines for solving complex video understanding problems.

\section{Related Work}
\label{sec:formatting}
\paragraph{Video LLMs}
Analyzing and understanding video content using large-scale language models (LLMs) typically involves fine-tuning or pre-training methods. Pre-training strategies, such as supervised or contrastive learning, develop video LLMs, while instruction fine-tuning updates adapter parameters to enable video comprehension \cite{tang2023video}. For example, LaViLa \cite{zhao2023learning} enhances video subtitle generation through a cross-attention module and rewriting mechanism, improving coverage and diversity. Video-LLaMA \cite{zhang2023video} addresses spatio-temporal visual variations using separate video and audio encoders with an advanced audio-visual Q-former, significantly boosting video comprehension. Video-LLaVA \cite{lin2023video} connects multimodal representations into a unified semantic space with LLMs, improving video understanding tasks.

However, these methods often consume a lot of computational resources and time during the training process, and the models can usually only target specific tasks related to the training data. In addition, video LLMs trained from scratch may not be able to achieve the expected performance when dealing with longer or previously unseen videos, showing shortcomings in understanding long videos and dealing with complex video question-answer tasks.

\paragraph{Long Video Understanding system with LLMs}

LLMs and Multimodal LLMs (MLLMs) applied to long video comprehension tasks utilize external systems to process extensive content. This involves analyzing visual elements, actions, scenes, and objects over time, aligning multimodal information with textual modalities, and leveraging the powerful text processing capabilities of MLLMs~\cite{tang2023video}. For instance, Vlog~\cite{vlog} uses pre-trained models for different modalities to record and interpret visual and audio information, summarizing it into detailed text for MLLM comprehension. MM-REACT \cite{yang2023mmreact} employs visual expert tools via internal prompts, enhancing MLLMs' visual understanding. MM-VID~\cite{lin2023mmvid} segments videos using ASR and Scene Detection tools, generating and integrating textual descriptions to complete Q\&A tasks with MLLMs. LLoVi \cite{zhang2023simple} uses a process that generates a summary based on subtitles and questions, then uses the summary for question-answering. VideoTree \cite{wang2024videotree} employs a three-step process to understand long videos, clustering video frames, calculating relevance, and performing depth expansion for question-answering. VideoAgent \cite{fan2024videoagent} preprocesses videos to generate captions and object data, and uses an agent with pre-provided tools to obtain answers.

Compared to training entirely new video LLMs, these MLLM-based approaches significantly reduce the need for computational resources while allowing the system to integrate or update external tools according to new technical or performance requirements. However, such long video understanding methodologies will lose a large amount of video information when transforming modalities, and do not fully utilize the multimodal processing capabilities of MLLM. In addition, these systems usually lack sufficient autonomy to support more complex video questioning and interaction, which limits their depth and breadth in practical applications.

\paragraph{MultiModal RAG}
MultiModal Retrieval Augmented Generation (RAG) leverages images, videos, audio, and other non-text data for information retrieval, enhancing content relevance and context for complex query and generation tasks~\cite{zhang2018multimodal}. The LlamaIndex \cite{Liu_LlamaIndex_2022} framework improves relevance and accuracy by enabling quick retrieval and processing of multimodal content through precise embedding and efficient indexing. Indexify \cite{indexify2024} provides a robust framework for building multimodal RAG systems with real-time data pipelines, extractor SDKs, and powerful storage interfaces for efficient information extraction and indexing. iRAG \cite{arefeen2024irag} uses AI model selection for perceptual queries, improving the speed and quality of multimodal data-to-text conversion, particularly for real-time, long video understanding.

Although these multimodal RAG systems offer great advantages in integrating multimodal information, they are still unable to eliminate the significant loss of information that occurs when data is transformed from video to knowledge in a RAG system. Our OmAgent, on the other hand, has no significant loss of information thanks for the task planning and autonomous tool call capability of DnC Loop especially the "rewinder" mechanism.

\section{Method}
\label{sec:method}
\begin{figure*}[h]
    \centering
    \includegraphics[width=1\textwidth]{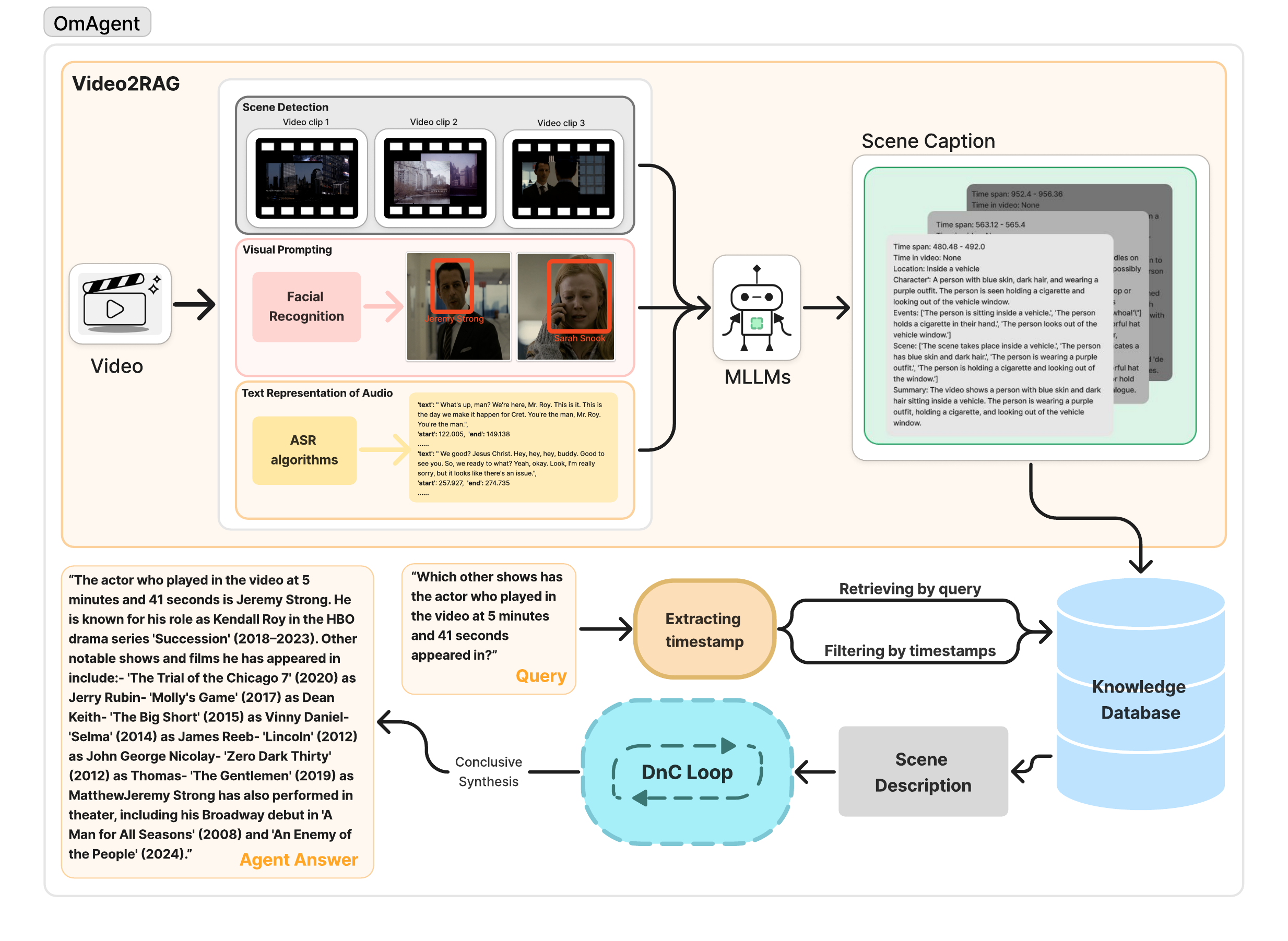}
    \caption{How OmAgent understand video. In Video2RAG, the video is processed by different algorithms (e.g. Scene Detection, ASR and face recognition) and then summarized by MLLMs to generate Scene Captions. Those captions are encoded and saved in the knowledge database. When OmAgent receives a query, it filters and retrieves in knowledge database based on timestamps (if available). The retrieved information is processed by the Divide-and-Conquer Loop and summarized by Conclusive Synthesis to generate the final answer.
}
    \label{fig1: OmAgent}
\end{figure*}

The process of OmAgent's video understanding can be bifurcated into two primary parts: Video2RAG and DnC Loop. As illustrated in Figure \ref{fig1: OmAgent}, all video data must undergo preliminary processing before being stored in the knowledge database in preparation for subsequent tasks. The preprocessing phase of Video2RAG encompasses a series of model identification and vectorization procedures, culminating in the extraction of the core content of the video files for storage.
When undertaking video understanding tasks, the initial step is to extract temporal information from the query. This information will then be used to filter the retrieved results. Subsequently, the query is encoded by text encoder, and the embedding is employed to retrieve pertinent video segment information from the knowledge database.  

The retrieved video clip information and the original task will be transmitted to DnC Loop, the intelligent agent capable of autonomously planning and executing tasks, for processing. Complex tasks will be recursively subdivided into executable subtasks. If at any point the agent deems that specific video details need to be reviewed, it will utilize the rewind tool to examine the relevant content. Once all subtasks are successfully completed, the execution results will be conveyed to a node dedicated to synthesizing the final answer.

\subsection{Video to RAG}
OmAgent's preprocessing (as shown in Figure \ref{fig1: OmAgent}) of video data is similar to a multimodal RAG. This approach avoids treating the entire content of a very long video as context input to the large language model, which would lead to three serious issues: (1) The length of the context would limit the maximum length of the video that can be processed. (2) Using an extremely long context for each question and answer session would cause an explosive increase in token usage. (3) An overly long context increases the difficulty of LLM inference, affecting the accuracy of question and answer sessions. OmAgent's Video2RAG processing mainly consists of the following steps.

\paragraph{Scene Detection}
Firstly, an algorithm is used to segment the video into relatively independent video blocks. The main purpose of this step is to locate the key nodes of the video. We can determine whether to segment the scene by assessing the degree of change in the frames; overly short segments will be merged together. The extracted video segments will have their start and end timestamps recorded, and 10 frames will be uniformly sampled from every segment.

\paragraph{Visual Prompting}
During the video preprocessing stage, additional algorithms can be used to provide more information. For example, using facial recognition, we can obtain information about the characters in the video. OmAgent will annotate this algorithmic information directly on the images through visual prompting, i.e., drawing the corresponding recognition boxes and using text to explain above the bounding box. This allows for the full utilization of the powerful understanding capabilities of MLLMs.
\paragraph{Text Representation of Audio}
The audio information in the video is as important as the visual information. OmAgent uses ASR algorithms to convert the speech in the video into text and employs speaker diarization algorithms to distinguish between different speakers.
\paragraph{Scene Caption}
Using MLLMs that support multiple images, each video segment's content is summarized. The inputs include video frames that have already been annotated with visual prompting and the transcribed audio information. In the process of generating dense captions at this step, We have delineated a set of pivotal elements to guide the MLLM in generating effective and comprehensive captions, ensuring that vital information is not overlooked in the absence of explicit objectives. OmAgent specifies the following dimensions as instructions to MLLMs:
\begin{itemize} [noitemsep, leftmargin=*, itemsep=0pt, topsep=0pt]
    \item The time information of the current video clip in terms of periods like morning or evening, seasons like spring or autumn, or specific years and time points. 
    \item Describe the location where the current event is taking place, including scene details.
    \item Provide a detailed description of the current characters, including their names, relationships, and what they are doing, etc.
    \item List all the detailed events in the video content in chronological order.
    \item Give some detailed description of the scene of the video. This includes, but is not limited to, scene information, textual information, character status expressions, and events displayed in the video.
    \item Provide an overall description and summary of the content of this video.
\end{itemize}
\paragraph{Encode and Save}
The final step in video processing involves vectorizing the scene captions and storing them in a vector database (knowledge database). Additionally, the original text of the captions is also stored in the memory for keyword-based retrieval. The start and end timestamps of the video segments are used as filtering fields and are likewise stored in the memory repository of the OmAgent agent.

\subsection{Divide-and-Conquer Loop}
In computer science, divide-and-conquer (DnC) is a highly classical algorithm design paradigm. A divide-and-conquer approach entails the iterative decomposition of a problem into multiple sub-problems. This process continues until the sub-problems reach a level of simplicity that allows for direct resolution~\cite{zhao2016dialport}. The solutions to these sub-problems are subsequently merged to yield a resolution to the initial problem. In order to ensure that OmAgent is not limited to simple video Q\&A functionality but possesses robust problem-solving capabilities, we initially aimed to build a general task-solving agent system when designing and constructing the agent framework. Inspired by XAgent’s ~\cite{xagent2023} double looped planner, we designed an agent framework based on the divide-and-conquer task processing loop (DnC Loop), which is capable of performing recursive task decomposition and execution. The DnC Loop task-solving procedure is shown in Algorithm 1.

\begin{algorithm}
\small
\caption{DnC Loop of OmAgent}
\begin{algorithmic}
\Require{The input Query $UserTask$; The max depth of the TaskTree $N$}
\State Initialize $Task$ = TaskTree.init($UserTask$)
\Procedure {DnC}{$Task, N$}
    \State $Result \gets \text{Conqueror}(Task)$
    \If {$Result.\text{type} = \text{``too complex''}$}
        \State $Subtasks \gets \text{Divider}(Task, Result.\text{reason})$
        \If {$Subtasks.\text{success}$}
            \State $Task$.add($Subtasks$.\text{tasks})
            \ForAll {$Subtask \in Task.\text{subtasks}$}
                \If {$Task.\text{depth} \leq N$}
                    \State \Call{DnC}{$Subtask, N$}
                \Else
                    \State \Return \text{``Task tree depth exceeded''}
                \EndIf
            \EndFor
        \Else
            \State \Return $Subtasks.\text{reason}$
        \EndIf
    \ElsIf {$Result.\text{type} = \text{``requires tool''}$}
        \State $ToolResult \gets \text{ToolCall}(Task, Result.\text{tool})$
        \State $Task.\text{update}(Task, ToolResult)$
        \State \Return $ToolResult$
    \ElsIf {$Result.\text{type} = \text{``direct answer''}$}
        \State $Task.\text{update}(Task, Result.\text{answer})$
        \State \Return $Result.\text{answer}$
    \EndIf
\EndProcedure
\end{algorithmic}
\end{algorithm}

\begin{figure*}[ht]
    \centering
    \includegraphics[width=1\textwidth]{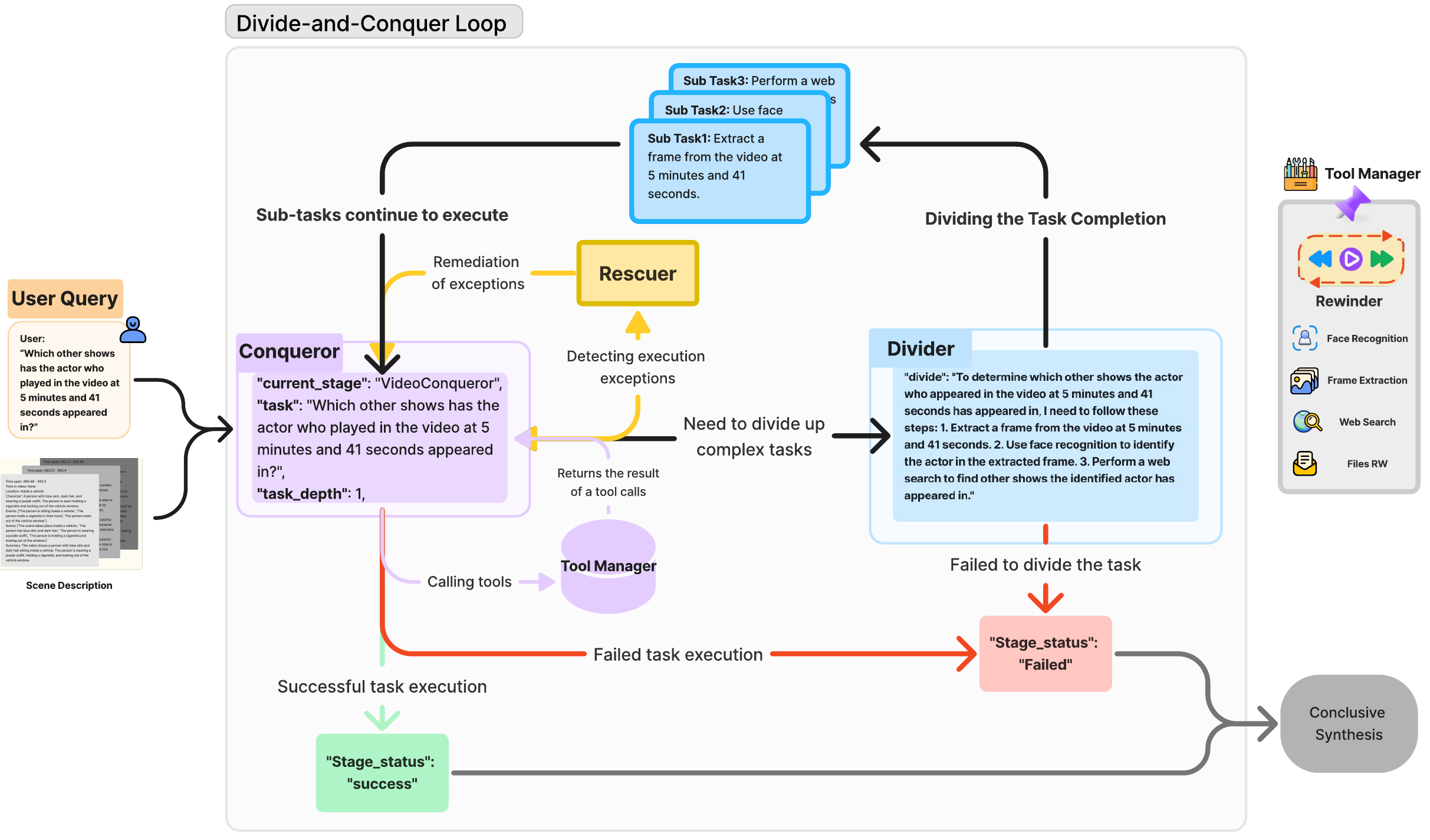}
    \caption{Divider and Conqueror Loop task-solving procedure. In the DnC Loop, simple problems are directly executed by Conqueror, while complex problems are split by Divider until they can be executed. The Rescuer recognizes exceptions and retries the task. The Tool Manager organizes the external tools. It is worth mentioning that the Rewinder tool can goes back through the entire video to find information and missing details. Finally, the DnC loop outputs the relevant content whether the execution fails or succeeds.
}
    \label{fig2: DNC}
\end{figure*}

\paragraph{Conqueror}
Conqueror is the entry point of the DnC loop. It is responsible for evaluating and processing the current task. For a given task, Conqueror may return one of the following three types of results:
\begin{itemize} [noitemsep, leftmargin=*, itemsep=0pt, topsep=0pt]
    \item If the current task is too complex and needs to be divided into subtasks, Conqueror returns the reason for the division.
    \item If the execution of the current task requires the use of a specific tool, Conqueror returns both the task information and the tool information. These pieces of information will be passed to the tool execution module for tool invocation.
    \item If the current task can be answered directly by the LLM, return the result directly.
\end{itemize}
Conqueror will detect the depth of the task tree and terminate task execution when it exceeds the user's setting to prevent tasks from being infinitely split. The position of Conqueror in the whole process is shown in Figure \ref{fig2: DNC}.
\paragraph{Divider}
The Divider component is responsible for breaking down complex tasks into simpler ones while ensuring that the execution results of these simple tasks are equivalent to the original task. When the Conqueror component determines the necessity of task division, it delegates the task to the Divider for attempted division. Successfully divided tasks are then integrated into the Task tree as child nodes of the original task node. If the division fails, the Divider is asked to provide the reason.
\paragraph{Rescuer}
Rescuer is an auxiliary module in the Conqueror's execution process. It attempts to repair issues and ensure the smooth completion of the Conqueror's execution when errors occur. A typical scenario is when the agent tries to execute a piece of code, but a required package is missing in the environment. The Rescuer can attempt to fix the runtime environment issue. The position of Divider in the whole process is shown in Figure \ref{fig2: DNC}.
\paragraph{Task tree}
In software development practice, divide-and-conquer is often implemented using recursion. OmAgent uses recursive tree structure to store all the paths of task execution. With the help of the Loop node, it achieves recursive operations for task decomposition and execution. 

\subsection{Tool call}
As a standard capability of intelligent agents, the principle of tool calling lies in utilizing the powerful logical generation ability of large language models (LLMs) to generate corresponding tool invocation request parameters based on task information. In addition to conventional tools, OmAgent specifically offers a video detail rewinder tool for further information extraction within specific time ranges of a video. OmAgent can autonomously choose to view details of a particular segment of a video when necessary, addressing the issue of information loss that occurs when video data transitions from a continuous information source to a discrete one during the preprocessing stage.

Furthermore, OmAgent provides conventional tools such as internet search tools, facial recognition tools, and file processing tools to meet more complex user tasks.

\section{Experimental Settings}
\label{sec:exp}
To validate the efficacy of the OmAgent system in addressing complex problems within real-world scenarios, we have designed a two-phase experimental approach:

\subsection{General problem-solving capabilities}
We hypothesize that understanding lengthy and intricate videos relies significantly on an agent's comprehensive problem-solving skills. To test this, we used two general-purpose intelligent benchmarks, MBPP~\cite{austin2021program} and FreshQA~\cite{vu2023freshllms}. We focused on the DnC Loop's ability to plan and execute tasks and its proficiency in utilizing tools to address complex issues
\paragraph{Datasets}


The Mostly Basic Programming Problems (MBPP) benchmark includes 976 elementary Python coding tasks. These problems are designed to evaluate the system's proficiency to plan solutions, select and invoke the right tools, and fix errors effectively.

FreshQA is a continuously updated collection of real-world questions and answers, reflecting the constantly changing nature of reality. As a result, FreshQA focuses on the system’s ability to learn and integrate new information from external sources.

\paragraph{Settings}
To study OmAgent's capability in comprehending complex long-form videos, we devised two control groups:

First, we aimed to ascertain the performance that could be achieved by one of the most advanced MLLMs - GPT-4o, based solely on a limited number of video frames (restricted to 20 by the Microsoft Azure GPT-4o service) and basic dialogue textual information. In this experiment, we initially extracted 20 frames evenly from the video and paired them with the dialogue text obtained through Whisper~\cite{radford2023robust} as the context for input into the MLLM for question-answering.

In the second control experiment, we sought to evaluate the strengths and weaknesses of using a multimodal RAG approach compared to our Agent strategy. To ensure a fair comparison, we isolated the Video2RAG component from OmAgent to serve as the RAG system. When a query is input, the system first retrieves relevant video clip information from the knowledge database, then inputs this pertinent data as context into the MLLM for question-answering.

\subsection{Long-form videos understanding capabilities}
We created a benchmark with over 2000 question-answer pairs to evaluate OmAgent's ability to understand, answer questions, and recall detailed information from long-form videos. We aimed to assess the general understanding of ultra-long video content and the ability to recall specific details. Our benchmark was designed to be logically coherent and narrative-rich, with video segments up to an hour long. This benchmark allows us to measure the capabilities of our intelligent agent and compare its performance against large language models.
\paragraph{Datasets}

Publicly available long-form video understanding datasets are very scarce, similar to the FreshQA/MovieQA datasets which only contain videos at the minute level, and the questions are not complex enough, for example, "How does Talia die?" in MovieQA, which is a question that can be inferred from consecutive frames, cannot meet our needs. SOK-Bench~\cite{wang2024sok} addresses scenarios slightly different from ours, as it demands the testing program to integrate situated and general knowledge to answer questions. MoVQA is a long movie question and answer dataset that utilizes 100 well-known movies to create complex long video question-answer pairs, which align with our requirements, but it is not yet open-sourced. Therefore, we created the dataset ourselves. We collected some long videos familiar to annotators, selected the top 100 videos in terms of frequency, and created 20 questions for each video. These questions were first proofread by two different annotators and then revised by a third annotator. Videos include episodes and movies, varieties, documentaries, and vlogs. These types of videos exhibit significant differences in themes, filming and editing techniques, data density, scene lengths, and alignment of video audio and visuals, fully demonstrating the diversity of the data.

\paragraph{Settings}
To evaluate the system's understanding of long-form videos and timelines, we defined questions in four categories: reasoning, information summary, event localization, and external knowledge. 

\begin{itemize} [noitemsep, leftmargin=*]
    \item \textbf{Reasoning} involves deducing relationships between events. 
    \item \textbf{Information summary} extracts key details from specific timestamps. 
    \item \textbf{Event localization} tests timeline accuracy. 
    \item \textbf{External knowledge} requires finding relevant but unmentioned information.
\end{itemize}

Reasoning, information summary, and external knowledge questions are multiple-choice. Event localization requires precise timestamps or time spans, with a deviation within ±2 seconds for timestamps and an IoU exceeding 90\% for time spans.

\section{Results and Analysis}
\label{sec:result}
\subsection{General problem-solving capabilities}
GPT-4, recognized as a benchmark for evaluating Large Language Models, is known for its strong reasoning abilities and serves as the baseline model for our experiments. XAgent, an advanced agent system, features a well-designed Dual-Loop Mechanism that allows it to address problems from both broad and detailed perspectives.
\begin{table}[h]
    \centering
    \begin{tabular}{ccc}
        \hline
         & \textbf{MBPP} & \textbf{FreshQA}\\
         \hline
        \textbf{GPT4+OmAgent} & \textbf{88.3\%} & \textbf{79.7\%}\\
        GPT4+XAgent & 84.2\% & 74.0\%\\
        GPT4 & 80.01\% & 67.0\%\\
        \hline
    \end{tabular}
    \caption{Results on MBPP and FreshQA comparing with GPT4 and XAgent. Showing OmAgent has a strong generalized task solving capbility.}
    \label{tab:table1}
\end{table}

\begin{table*}[h]
\centering
{%
\begin{tabular}{llllll}
\hline
\textbf{}            & \textbf{Vlog} & \textbf{Episode and Movies} & \textbf{Variety} & \textbf{Documentary} & \textbf{Total} \\
\hline
\textbf{OmAgent}       & \textbf{57.14\%}        & \textbf{56.25\%}            & \textbf{23.53\%}            & \textbf{36.84\%}       & \textbf{45.45\%}        \\
Video2RAG & 42.86\%         & 32.35\%            & 19.88\%             & 31.57\%      & 27.27\%          \\
Frames with STT  & 42.85\%         & 29.41\%            & 17.64\%            & 31.58\%      & 28.57\%           \\
VideoAgent   & 41.72\% & 23.53\%             & 11.76\%  & 26.32\%      & 23.38\%  \\
VideoTree & 34.52\% & 31.48\% & 21.27\% & 27.35\%   & 26.76\% \\
LLoVi & 28.57\% & 24.16\% & 17.65\% & 21.05\% & 23.63\% \\
\hline
\end{tabular}%
}
\caption{Results of different types of videos on OmAgent and the other five baselines. }
\label{tab:table2}
\end{table*}

\begin{table*}[h]
\centering
\resizebox{\textwidth}{!}{%
\begin{tabular}{lllll}
\hline
\textbf{}            & \textbf{Reasoning} & \textbf{Event Localization} & \textbf{Information Summary} & \textbf{External Knowledge} \\
\hline
\textbf{OmAgent}       & \textbf{81.82\%}        & \textbf{19.05\%}            & \textbf{72.74\%}            & \textbf{57.21\%}       \\
Video2RAG & 72.73\%         & 4.76\%            & 50.17\%             & 23.36\%      \\
Frames with STT  & 63.64\%         & 2.38\%            & 63.63\%            & 19.46\%      \\
VideoAgent & 64.66\% & 2.25\% & 45.45\% & 23.78\%   \\
VideoTree & 35.30\% & 18.62\% & 47.27\% & 29.57\%   \\
LLoVi & 27.27\% & 11.90\% & 45.46\% & 24.57\% \\
\hline
\end{tabular}%
}
\caption{Results of different types of queries on OmAgent and the other five baselines.}
\label{tab:table3}
\end{table*}

\subsection{Long-form videos understanding capbilities}
The results in Table \ref{tab:table1} clearly show that both agent systems outperform the basic inferential capabilities of GPT-4 alone. Notably, OmAgent surpasses XAgent~\cite{xagent2023} in overall performance. Analysis reveals that XAgent's Dual-Loop Mechanism, while thorough, often leads to overthinking and complicates problem-solving. In contrast, OmAgent's Rescuer mechanism proves more effective, especially in handling code-related tasks. This mechanism enables OmAgent to dynamically correct issues based on real-time results, leading to superior performance.
Table \ref{tab:table2} compares the scores of five baselines and OmAgent across different types of long-form video understanding. OmAgent achieved the highest scores. Vlogs, variety, and documentaries contain extensive narration, so the STT data and the resulting scene captions encompass most relevant information of the video. Therefore, the performance difference between OmAgent and the other two methods in these categories is not as significant as in episodes and movies. In episodes and movies, scenes change frequently, complex queries involve cross-scene information, and STT data might span scene transitions. Compared to frames with STT, Video2RAG retrieves data related to the query, reducing data redundancy, and thus has higher scores than frames with STT. However, since it only retrieves relevant information from a vector database, complex questions such as "Are there any scene changes between 03:58 and 04:02, and what is their connection?" are not achieved. On the other hand, OmAgent’s DnC Loop breaks down complex questions into several sub-questions, including "Extract frames between 03:58 and 04:02," "Analyze the extracted frames to identify any scene changes," and "Determine the connection between the scenes based on the identified changes." By leveraging the rewinder capability, it pinpoints the relevant segments for rewatching, thereby arriving at the correct answer.

Furthermore, we conducted a detailed analysis of different question types. Table \ref{tab:table3} provides a comparison of OmAgent and five baselines in terms of reasoning, event localization, information summary, and external knowledge. The results show that OmAgent achieves the highest scores in all four types of questions. Through its rewinder capability, OmAgent can extract more detailed video information and accurately locate timestamps, which leads to significant improvements in reasoning, event localization, and information summarization tasks compared to frames with STT and Video2RAG. External knowledge task has stricter requirements for information retrieval. Although GPT can answer some questions through its own capabilities and scene information, OmAgent achieves higher scores by utilizing various external tools (such as facial recognition, web search, etc.) to obtain more accurate relevant information. Notably, in the question type of information summary, Video2RAG scored lower than frames with STT. Analysis reveal that video2RAG's information source came from scene captions generated based on STT and images within the scene, which had a certain probability of omissions compared to the original STT, thus affecting the scores. The higher frame extraction frequency of VideoAgent, VideoTree and LLoVi allowed them to capture more visual information, but their lack of audio processing led to suboptimal performance in many scenarios. VideoAgent’s lack of DnC and rewinder capabilities limited its effectiveness in query types requiring detailed information comparing with OmAgent. Although OmAgent performed the best across all question types, it still scores relatively low in the event localization task. Detailed study and analysis reveal that LLM tends to directly use timestamps or time spans within the scene, leading to a lack of precision in answers. OmAgent's DnC loop and rewinder capabilities can mitigate this issue but cannot completely resolve it.

\section{Conclusion}
\label{sec:conclusion}
OmAgent is a powerful video comprehension agent that integrates multimodal RAG with a generalist AI agent, enabling several advanced capabilities. It offers a theoretical near-infinite length video understanding capacity and incorporates a secondary recall mechanism for detailed video information, which significantly mitigates information loss. Additionally, OmAgent autonomously invokes tools based on video comprehension tasks, allowing it to execute more intricate operations. These capabilities have resulted in remarkable performance on the MBPP, FreshQA, and our proposed long video complex task test datasets.

\section{Limitations}
\label{sec:limitations}
\begin{itemize}
    \item When determining the positioning of an event in a video, LLM tends to directly use the timestamps given in the scene. OmAgent can alleviate this situation through DnC loop and rewinder but cannot completely resolve it.
    \item The character information in long-form videos is usually diverse. For example, character Logan can be both a boss and a father, making it difficult to align back to the same person, causing OmAgent to fail in precise positioning. It is necessary to add a visual prompt to the character for suppression.
    \item The phenomenon of audio-visual asynchrony is prominent in long-form videos. For example, in documentaries, the picture might not change significantly while the speech has already shifted to introduce other scenes, resulting in a misalignment between the picture and the speech. Additionally, STT only processes the speech of characters, losing other audio information such as background music, sound effects, and different characters' voiceprints.
\end{itemize}

\section{Acknowledgements}
This research is partially supported by National Key R\&D Program of China under grant No. 2022YFF0902600

\bibliography{custom}

\appendix
\section{Case Study}
\label{sec:appendix}
In this section, we will illustrate the process by which OmAgent addresses complex video understanding problems through several examples. The primary aspects showcased include: scene caption, inner steps, and agent output. The scene caption displays sample data stored in the knowledge database after the Video2RAG preprocessing; the inner steps reveal the task decomposition and execution within the DnC Loop; the agent output presents the final results produced by OmAgent.

\subsection{Case 1}
\begin{figure}[h]
    \centering
    \includegraphics[width=0.5\textwidth]{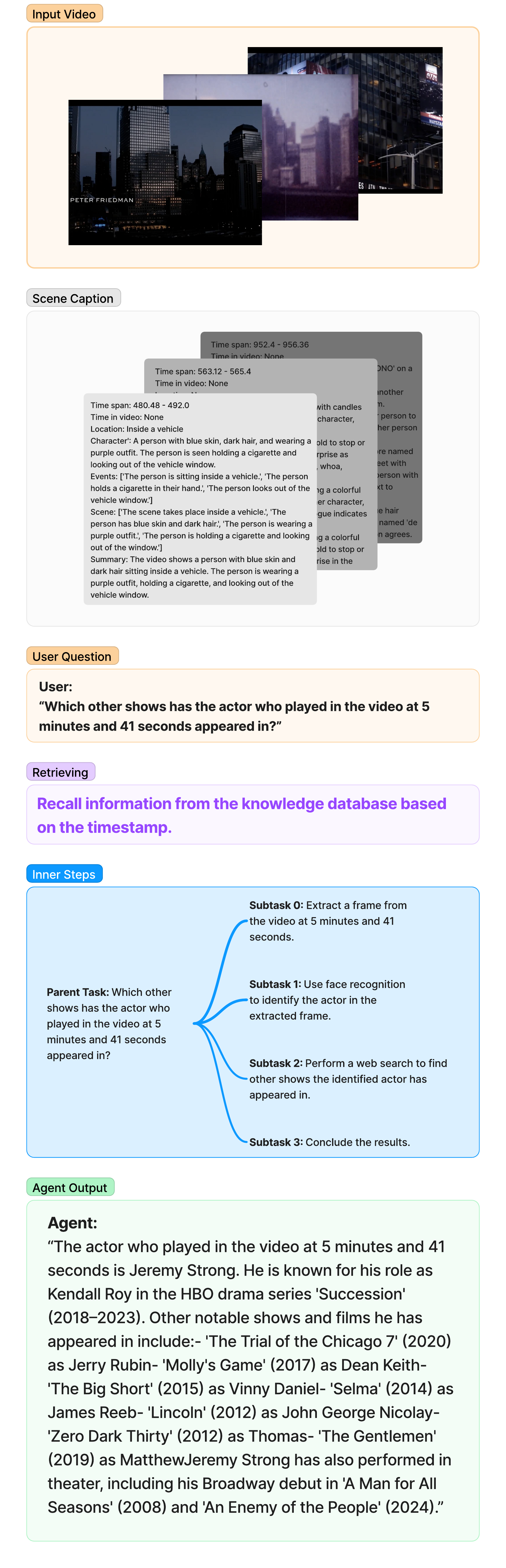}
    \caption{case 1}
    \label{apxfig1: case1}
\end{figure}

In this case in Figure \ref{apxfig1: case1}, the video we utilized is from the series "Succession," Season 1, Episode 1. The question posed was, "Which other shows has the actor who played in the video at 5 minutes and 41 seconds appeared in?" This is an intricate query, as the system must first identify the individual at the specified timestamp and subsequently gather information about the actor from the internet. Observably, the intelligent system deconstructed the task into three sub-tasks and then sequentially employed relevant tools to address each one, ultimately arriving at the correct answer.

\subsection{Case 2}
\begin{figure}[h]
    \centering
    \includegraphics[width=0.5\textwidth]{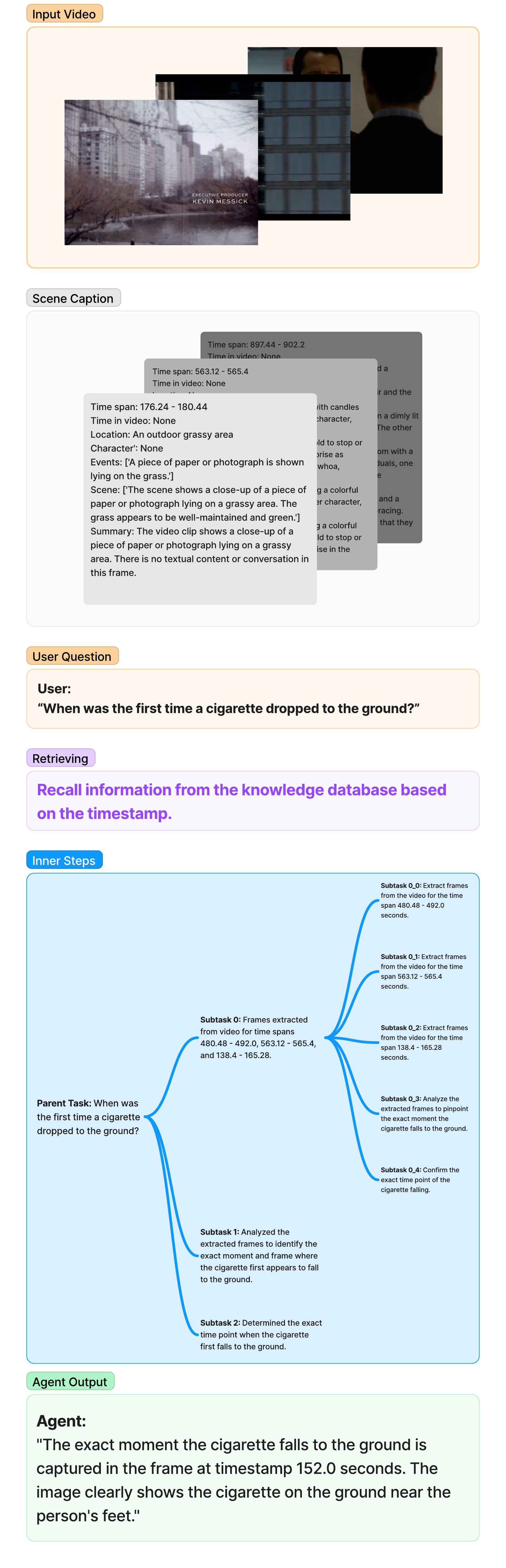}
    \caption{case 2}
    \label{apxfig1: case2}
\end{figure}

In this case as showing in Figure \ref{apxfig1: case2}, the video we utilized is from the series "Succession," Season 1, Episode 1. The question posed was, "When was the first time a cigarette dropped to the ground?" This task requires OmAgent to locate a highly detailed scene within a lengthy video. OmAgent initially divides this problem into three subtasks. Among these, the task of identifying the key frame is too complex to be completed directly and is further divided into four subtasks. It necessitates extracting details and summarizing results from three segments of the video. Ultimately, all the subtasks are successfully handled, and OmAgent returns the correct answer.

\section{Sample Experiment Results}
Tables \ref{tab:exp1}, \ref{tab:exp2}, and \ref{tab:exp3} show 20 examples of OmAgent’s performance on 2000 long-form video understanding questions. In these tables, "Video" refers to the video names, and "Question" refers to the questions in the dataset, with the choice options directly following the questions. OmAgent represents our answer output, while Video2RAG and frames with stt are the outputs of two control groups.
\begin{table*}[]
\centering
\resizebox{\textwidth}{!}{%
\begin{tabular}{p{2cm}|p{7cm}|p{2cm}|p{2cm}|p{2.5cm}|p{2cm}|p{2cm}}
\textbf{Video} & \textbf{Question} & \textbf{Question Type} & \textbf{Ground Truth} & \textbf{OmAgent} & \textbf{Video2RAG} & \textbf{Frame with STT} \\
\hline
Succession S01EP01 & When was the first time a cigarette dropped to the ground? & Event Localization & 00:02:32 & 00:02:32 & {[}00:02:18, 00:02:45{]} & Fail to answer \\
\hline
Succession S01EP01 & The task was to determine if the video shows a theme park? Please choose from the following options: a. The playground's interior scenes appeared between 08:58 and 09:49. b. The playground's interior scenes appeared after 09:50. c. The playground's interior scenes were shown before 08:57. d. No playground scenes appeared during this time. & Reasoning & a & a & a & None of a, b, c, d \\
\hline
Succession S01EP01 & In which time period does the headphone first appear? & Event Localization & {[}00:01:46, 00:01:49{]} & {[}00:01:46, 00:01:49{]} & {[}00:04:15, 00:04:27{]} & 0:01:46 \\
\hline
Succession S01EP01 & What important event occurs between the 34th and 35th minute? Choose from the following options: a. Connor gets angry in the bathroom, breaks the mirror and sink. b. Connor gets angry in the bathroom, vents his anger by smashing a vase, shouting loudly, etc. c. Connor gets angry in the bathroom, kicks over the toilet and bathtub. d. Connor gets angry in the bathroom, smashes the door and windows. & Reasoning & b & b & b & b \\
\hline
Westworld S01EP01 & Please summarize the content of this video. Choose from the following options: a. The video starts with a dialogue in the protagonist's dream, introducing a Western world live-action game built by modern people that attracts various customers to experience. The protagonist, as the main NPC, repeats the same life continuously. b. After an update in the Western world game, NPCs start experiencing various minor bugs. This puzzles the programmers behind the scenes, who begin inspecting the problematic NPCs. Initially, only some NPCs experience issues like lag during operation, which the programmers do not pay much attention to after updating them. c. The protagonist's father awakens to consciousness after seeing a photo of the modern world and recites a line from a Shakespearean work to the protagonist. This triggers uncontrollable events in the entire game world. d. The administrator questions both the protagonist and her father. The protagonist enters a state of consciousness during the conversation and claims to retaliate against the boss. The dialogue with the protagonist echoes the dream at the beginning, foreshadowing the occurrence of certain events. & information summary & a, b, c & b & a & a \\
\hline
Westworld S01EP01 & Please find the specific time when the female protagonist wakes up from the bed. & Event Localization & 00:02:46, 00:14:31, 00:44:28, 01:04:15 & 00:44:23 & {[}00:00:00, 00:00:04{]} & {[}00:14:26, 00:14:32{]} \\
\hline
Westworld S01EP01 & From 01:04:03 to 01:05:59, what changes occurred in the position of the female protagonist, and what do these changes indicate? Choose from the following options: a. During this time, the female protagonist went to her home in Westworld but did not return to the modern laboratory, indicating that Westworld is real. b. During this period, the female protagonist's movements from the modern laboratory to Westworld and then to her home reveal the authenticity of her life. c. During this period, the female protagonist's movements from the laboratory to Westworld and then to her home imply that Westworld and her home are real. d. In this time frame, the female protagonist's movements from the modern laboratory to the home in Westworld and back to the modern laboratory indicate the falsity of Westworld. & Reasoning & d & d & d & d
\end{tabular}%
}
\caption{OmAgent Experiment Results}
\label{tab:exp1}
\end{table*}

\begin{table*}[]
\centering
\resizebox{\textwidth}{!}{%
\begin{tabular}{p{2cm}|p{7cm}|p{2cm}|p{2cm}|p{2.5cm}|p{2cm}|p{2cm}}
\textbf{Video} & \textbf{Question} & \textbf{Question Type} & \textbf{Ground Truth} & \textbf{OmAgent} & \textbf{Video2RAG} & \textbf{Frame with STT} \\
\hline
Westworld S01EP01 & Why was Dolores' father being inpsected? Choose from the following options, it's a multiple choices question: a. Because his program malfunctioned and he couldn't continue to function properly. b. Because he displayed emotions that a host NPC shouldn't have. c. Because he was killed by bandits in the mountains and needed servicing to be restored. d. Dolores' father was serviced because he showed strong motives and drives when questioned. & Reasoning & b, d & b, d & a & a \\
\hline
Tempting Fortune S01EP01 & In the video, when was the first money spent and what was it spent on? & Event Localization & 00:31:14 & {[}00:31:08, 00:31:11{]} & {[}00:04:14, 00:04:26{]} & {[}00:05:51, 00:05:57{]} \\
\hline
Tempting Fortune S01EP01 & Please describe the scene at 42:40. Select from the following options: a. A group of people are gathered around a hearth roasting food, with several lamps extinguished nearby. b. A group of people are sitting around a campfire, enjoying dinner, with many lights illuminated nearby. c. A large crowd is dining around a candelabrum, with many lamps turned off nearby. d. Some people are singing around a bonfire, with dim lights in the vicinity. & information summary & b & b & None of a, b, c, d & b \\
\hline
Tempting Fortune S01EP01 & At what time in the video does someone fall into the river? & Event Localization & 00:45:56 & 00:45:54 & None of a, b, c, d & {[}00:29:38, 00:30:17{]} \\
\hline
Tempting Fortune S01EP01 & At what time did all team members climb up the hill? & Event Localization & {[}00:12:45, 00:12:51{]} & {[}00:12:37, 00:12:51{]} & None of a, b, c, d & None of a, b, c, d \\
\hline
A Perfect Planet & When do volcanic eruptions occur in the video? & Event Localization & {[}00:01:37, 00:03:55{]}, {[}00:14:30, 00:15:00{]}, {[}00:43:24, 00:44:05{]} & {[}0:01:36, 0:01:41{]}, {[}0:03:03, 0:03:06{]}, {[}0:03:21, 0:03:28{]}, {[}0:03:28, 0:03:37{]}, {[}0:03:45, 0:03:55{]}, {[}0:14:30, 0:14:37{]}, {[}0:14:37, 0:15:00{]}, {[}0:17:15, 0:18:09{]}, {[}0:43:24, 0:43:36{]}, {[}0:43:36, 0:43:50{]}, {[}0:43:58, 0:44:05{]} & {[}0:01:36, 0:01:41{]}, {[}0:01:52, 0:02:15{]}, {[}0:02:40, 0:02:53{]}, {[}0:14:30, 0:14:37{]}, {[}0:43:24, 0:43:36{]}, {[}0:43:36, 0:43:50{]}, {[}0:43:58, 0:44:05{]} & {[}0:02:40, 0:02:53{]}, {[}0:03:03, 0:03:06{]}, {[}0:04:04, 0:04:10{]}, {[}0:15:00, 0:15:10{]}, {[}0:15:40, 0:15:57{]}, {[}0:18:09, 0:19:10{]}, {[}0:21:23, 0:21:31{]}, {[}0:26:19, 0:26:43{]}, {[}0:31:37, 0:31:38{]}, {[}0:38:18, 0:38:32{]}, {[}0:43:07, 0:43:16{]}, {[}0:48:04, 0:48:34{]} \\
\hline
A Perfect Planet & What happened between 10:38 and 10:53? Choose from the following options: a. A little chick couldn't overcome the hardships on the way and finally collapsed exhausted in a mud pit. b. A little chick encountered many difficulties on the road and eventually collapsed on the grass from exhaustion. c. A bird encountered many obstacles during flight and eventually collapsed in the desert. d. A little chick went through various hardships and finally collapsed exhausted on a rock. & information summary & a & a & b & a \\
\hline
A Perfect Planet & When did the Galapagos land iguana appear in the video? & Event Localization & {[}00:15:58, 00:21:47{]} & {[}0:21:31, 0:21:42{]} & {[}0:15:00, 0:21:31{]} & {[}0:15:57, 0:16:47{]} \\
\hline
A Perfect Planet & When did the East African Rift Valley appear in the video? & Event Localization & {[}00:44:12{]} & {[}0:44:05, 0:44:18{]} & {[}0:44:05, 0:44:18{]} & {[}0:44:05, 0:45:59{]}
\end{tabular}%
}
\caption{OmAgent Experiment Results}
\label{tab:exp2}
\end{table*}

\begin{table*}[]
\centering
\resizebox{\textwidth}{!}{%
\begin{tabular}{p{2cm}|p{7cm}|p{2cm}|p{2cm}|p{2.5cm}|p{2cm}|p{2cm}}
\textbf{Video} & \textbf{Question} & \textbf{Question Type} & \textbf{Ground Truth} & \textbf{OmAgent} & \textbf{Video2RAG} & \textbf{Frame with STT} \\
\hline
WWDC 2024 Recap: Is Apple Intelligence Legit? & What new features mentioned in the video have Android users long been able to enjoy? a. The ability to freely arrange and customize home screen icons. b. Cleanup tool in the photo app. c. Remote control of other people's devices. d. Hide apps. & Reasoning & a, b & a, b, d & a, d & a, b, c, d \\
\hline
WWDC 2024 Recap: Is Apple Intelligence Legit? & What is the author's attitude towards the iPad OS update? a. Disappointing. b. Exciting. c. There are some great upgrades, but not quite enough. d. Not mentioned. & Reasoning & c & c & c & c \\
\hline
WWDC 2024 Recap: Is Apple Intelligence Legit? & At 14 minutes and 26 seconds, what address is displayed on the phone? a. Waterbar. b. San Francisco Fisherman's Wharf. c. San Francisco International Airport. d. Ronald Reagan Airport. & information summary & c & c & b & b \\
\hline
WWDC 2024 Recap: Is Apple Intelligence Legit? & At what time in the video does the host promote their channel? & Event Localization & {[}0:05:54, 0:06:03{]} & {[}0:05:24, 0:06:03{]} & {[}0:06:03, 0:06:09{]} & {[}0:06:03, 0:06:09{]}

\end{tabular}%
}
\caption{OmAgent Experiment Results}
\label{tab:exp3}
\end{table*}
\end{document}